\pdfoutput=1

\documentclass[11pt]{article}

\usepackage{url}
\usepackage{multirow}
\usepackage{array}
\usepackage{graphicx}
\usepackage{algorithm} 
\usepackage{algpseudocode}
\usepackage{hyperref}

\usepackage{naacl2021}

\usepackage{times}
\usepackage{latexsym}
\usepackage{amsmath}

\usepackage[T1]{fontenc}

\usepackage[utf8]{inputenc}

\usepackage{microtype}

\title{Unsupervised Contextual Paraphrase Generation using Lexical Control and Reinforcement Learning}

\author{Sonal Garg$^{*}$ \\
  IIIT Bangalore \\
  \texttt{\small sonal.garg@iiitb.ac.in} \\\And
  Sumanth Prabhu$^{*}$ \\
  Applied Research, Swiggy \\
  \texttt{\small sumanth.prabhu@swiggy.in} \\\And
  Hemant Misra \\
  Applied Research, Swiggy \\
  \texttt{\small hemant.misra@swiggy.in} \\\AND
  G. Srinivasaraghavan \\
  IIIT Bangalore \\
  \texttt{\small gsr@iiitb.ac.in} \\
  }

\date{}

\begin{document}
\maketitle
\renewcommand{\thefootnote}{\fnsymbol{footnote}}
\footnotetext[1]{Equal contribution.}

\begin{abstract}

Customer support via chat requires agents to resolve customer queries with minimum wait time
and maximum customer satisfaction. Given that the agents as well as the customers can have varying levels of literacy, the overall quality of responses provided by the agents tend to be poor if they are not predefined. But using only static responses can lead to customer detraction as the customers tend to feel that they are no longer interacting with a human. Hence, it is vital to have variations of the static responses to reduce monotonicity of the responses. However, maintaining a list of such variations can be expensive. Given the conversation context and the agent response, we propose an unsupervised framework to generate contextual paraphrases using autoregressive models. We also propose an automated metric based on {\em Semantic Similarity}, {\em Textual Entailment}, {\em Expression Diversity} and {\em Fluency} to evaluate the quality of contextual paraphrases and demonstrate performance improvement with Reinforcement Learning (RL) fine-tuning using the automated metric as the reward function.



\end{abstract}

\section{Introduction}

Paraphrase generation is the task of generating text sequences that are candidate paraphrases of a reference text sequence. A text paraphrase essentially expresses the information contained in the text without using all the words present in it. Concretely, two text sequences are paraphrases of each other if they have high semantic similarity and expression diversity. Paraphrase generation finds numerous applications in Natural Language Understanding (NLU) tasks including question answering~\cite{QANet-ICLR-2018}, machine translation~\cite{thompson-post-2020-automatic}, text summarization~\cite{10.5555/3298023.3298027} and data augmentation for dialogue systems~\cite{hou2018sequencetosequence}. 

In this paper, we consider the use-case of assisting agents during customer-agent chat conversations where customers reach out to customer support for a problem and agents help resolve their issues. Often the overall quality of the responses provided by the agents tend to be poor, unless they are not predefined, owing to the widely varying levels of communication skills among the agents along with their tendency to give a safe `canned' response. Hence, customer support agents typically pre-define a set of candidate responses for all possible scenarios, with the goal of ensuring quicker resolutions to customer queries. However, using only static responses can affect the customer experience where customers feel that they are no longer interacting with a human but are exposed to a fixed decision tree. Moreover, customers have a tendency to be ungrammatical with spelling and inflectional variations while conversing with the agents using free text which can further affect the response quality. Hence, it is vital to have variations of the static responses to ensure a positive customer experience. Given a conversation context and a candidate response, we propose to build a system that generates candidate paraphrases that are more contextual and `informal' without conflicting the pre-defined responses mandated for a scenario, to ensure a positive customer experience. This is a unique problem setting under paraphrase generation where the generated paraphrases are required to not just literally paraphrase the actual agent response but also account for the context of the conversation.

Traditionally, paraphrase generation relies on the availability of labelled paired text sequences. However, collecting this training data can be very expensive. Popular paraphrase datasets like Quora Question Pairs~\footnote[2]{https://www.kaggle.com/c/quora-question-pairs}, ParaNMT
~\cite{wieting-gimpel-2018-paranmt}, PAWS~\cite{48578}, PAWS-X~\cite{yang2019pawsx} and MSCOCO~\cite{lin2014microsoft} are not extensible to our problem setting which demands handling domain specific ungrammatical customer queries and agent responses, and also requires the paraphrases to incorporate the context of the conversation. Several recent works \cite{Miao2019CGMHCS,liu-etal-2020-unsupervised,niu2020unsupervised} achieved state-of-the-art results in paraphrase generation using unsupervised settings. Inspired by these results, we explore unsupervised approaches to generating paraphrases for agent responses in chat conversations.

Wu et al. \cite{wu2020controllable} demonstrated that controlled text generation~\cite{hokamp-liu-2017-lexically, keskar2019ctrl, tang2019targetguided} can improve the quality of the results in conversational response generation. We extend the concept to our problem setting where contexts combined with control words identified from the agent responses are leveraged to generate candidate paraphrases of the responses. To evaluate the overall quality of the generated paraphrases, we use multiple automated metrics each of which  quantifies a different aspect of the response and finally combine them into an overall composite evaluation metric for the paraphrase. Specifically, we measure the degree of {\em textual entailment} of the generated response from the context, the {\em semantic similarity} and {\em expression diversity} of the generated response with respect to the actual agent response and the {\em language fluency} of the generated response. Concretely, we train a custom model to evaluate the textual entailment while leveraging BERTScore~\cite{bert-score} to quantify semantic similarity, a BLEU~\cite{bleu} based metric to quantify expression diversity and train another custom model to quantify the fluency. We then build a framework based on Reinforcement Learning (RL) that uses a combination of these individual automated metrics as the reward function.

The main contributions of our paper are as follows :
\begin{enumerate}
\item We propose an unsupervised framework for the task of paraphrase generation of agent responses in customer-agent chat conversations.
\item We propose a composite automated paraphrase quality evaluation metric based on individual Semantic Similarity, Textual Entailment, Expression Diversity and Fluency scores to evaluate contextual paraphrases of chat conversations that are not necessarily grammatically correct.
\item We propose a RL framework initialized with the model trained in an unsupervised fashion and fine-tuned using the automated metric as the reward. 
\end{enumerate}
 
Rest of the paper is organized as follows: In Section~\ref{RelatedWork}, we describe previous work for solving similar problems. Section~\ref{Methodology} describes our approach to contextual paraphrase generation in detail followed by Section~\ref{Experiments} where we explain the experiment settings to validate our approaches with the corresponding results. Finally, we present our findings and describe future work in Section~\ref{Conclusion}

\section{Related Work}
\label{RelatedWork}
Paraphrase generation is an important problem in NLU, especially in dialog systems. Typically, the task of generating paraphrases has been formalized as a sequence-to-sequence (Seq2Seq) learning problem. Prakash et al. \cite{prakash-etal-2016-neural}
employed a stacked residual long short-term memory (LSTM) network in the Seq2Seq model. They show that addition of residual connections enhances the model capacity to make use of deep reinforcement learning for paraphrase generation in a supervised fashion. Gupta et al.  \cite{gupta2018} explored combination of deep generative models with LSTM models to generate paraphrases.
Ma et al. \cite{ma-etal-2018-query} proposed to generate the words by querying distributed word representations using attention. Kazemnejad et al. \cite{kazemnejad-etal-2020-paraphrase} took a retrieval based approach where pairs of candidate paraphrases were extracted from a predefined index to train a model to generate paraphrases. Our problem setting does not benefit from these solutions due to the unavailability of training data in the form of paired contextualized paraphrases.

\subsection{Unsupervised Paraphrase Generation}
Paraphrasing using unsupervised approaches has been explored over several years. Earlier approaches~\cite{barzilay-lee-2003-learning,zhang-patrick-2005-paraphrase,qui-emnlp-2006} relied on lexical matching to identify paraphrases. More recent approaches leverage Variational Autoencoders (VAE)~\cite{gupta2018, Siddique_2020} that have been popular in text generation based applications and can be intuitively extended to paraphrase generation in an unsupervised fashion~\cite{bowman-etal-2016-generating, zhang-etal-2019-syntax}. Miao et al. \cite{Miao2019CGMHCS} showed that VAE based approaches suffer from error accumulation during the decoding phase and addressed sentence generation by directly sampling from the sentence space using Metropolis Hastings (MH) algorithm~\cite{metropolis1953equation}. However, the generated paraphrases from this approach lacked semantic similarity to the input~\cite{Siddique_2020}. Liu et al. \cite{liu-etal-2020-unsupervised} cast the paraphrase generation as an optimization problem, where they searched the sentence space for candidate paraphrases based on language fluency, expression diversity and semantic similarity. As pointed out by~\cite{Siddique_2020}, they did not effectively explore the entire sentence space, resulting in paraphrases that were not different enough from the input. Niu et al. \cite{niu2020unsupervised} relied on pre-trained autoregressive models to generate text while modifying the decoding algorithm to ensure high quality paraphrase generation. Our approach is based on reinforcement learning where the quality of the paraphrase is ensured using a reward combining textual entailment, semantic similarity, expression diversity and fluency. 

\subsection{Text Generation using Lexical control}
Controllable text generation has been an active area of research~\cite{hokamp-liu-2017-lexically, keskar2019ctrl, tang2019targetguided} where neural conversation models are typically guided towards a task specific targeted output. Wu et al. \cite{wu2020controllable} modeled conversational response generation by combining lexical control with external knowledge bases to further improve the relevance of the generated text. We extend these frameworks to paraphrase generation in conversations where the quality of paraphrases is improved by applying lexical control to identify the important tokens in the agent response and also including the context of the conversation to ensure relevance of the generated paraphrases. 

\subsection{Reinforcement Learning}
Deep reinforcement learning (DRL) for natural language processing has become an active area of research. In the last few years, we have seen great advancement in the number of applications and methods of RL. Li et al. \cite{DBLP:journals/corr/abs-1711-00279} built a paraphrase generation model based on a generator-evaluator reinforcement learning framework  using an LSTM based generator and inverse reinforcement learning based evaluator. Unlike our problem setting, the approach still needed training data in the form of paired paraphrases to train the evaluator in both online and offline versions. Ranzato et al. \cite{ranzato2015sequence} directly optimized the lexical based metric which is used during test time such as BLEU~\cite{papineni2002bleu} and ROUGE~\cite{lin2004rouge} as a reward function. However these algorithmically defined lexical measures loosely match with paraphrase generation factors \cite{liu-etal-2010-pem}. We extend our unsupervised model to train a policy via  Proximal Policy Optimization
(PPO)~\cite{schulman2017proximal} where the reward is calculated by an automated metric as in~\cite{Siddique_2020}. Our reward scheme however additionally includes textual entailment (that is extremely important in a chat-conversation) to ensure the generated paraphrase also uses  the given context appropriately.


\section{Methodology}
\label{Methodology}
\begin{figure}
  \includegraphics[width=7cm]{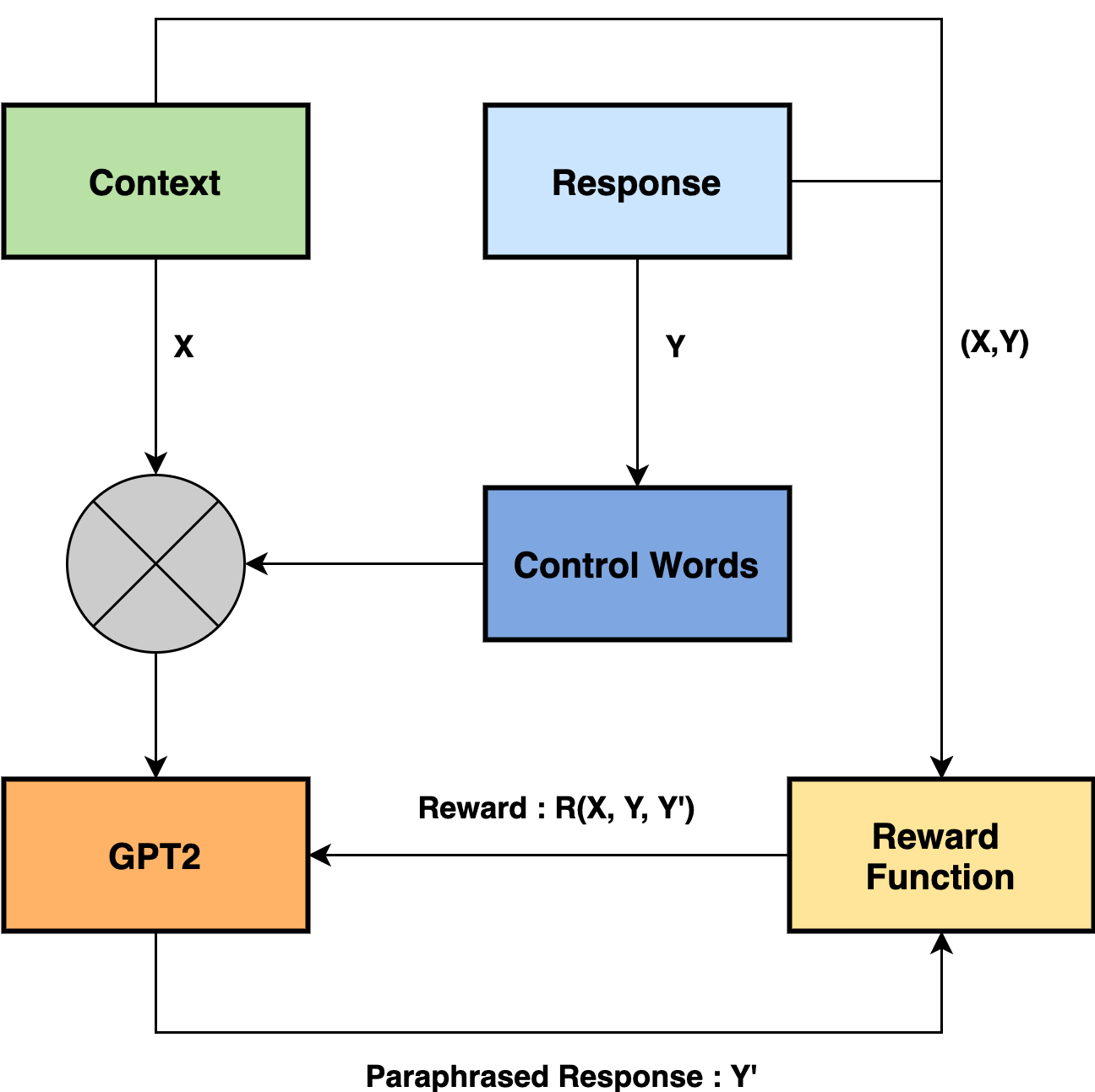}
  \caption{High level overview of the steps involved in RL fine-tuning of GPT-2 for Contextual Paraphrase Generation}
\end{figure}
  In this section, we describe our approach and the different components of our automated evaluation metric. We then further improve the approach via reinforcement learning (RL).

\subsection{Lexical Control with autoregressive language models}\label{lexical-control-methodology}
Large scale pre-trained autoregressive language models like GPT-2~\cite{radford2019language} have proven to perform well with multiple NLU tasks~\cite{radford2019language}. Although such models generate highly coherent text, they fail to generate relevant text without domain specific fine-tuning. 

Traditionally fine-tuning for paraphrase generation demands the availability of training data for paraphrases. In this paper, we propose an unsupervised framework for paraphrase generation of agent responses in customer-agent chat conversations. We generate paraphrases by fine-tuning GPT-2 with lexically controlled agent responses in combination with the context of the conversation.

Our framework does not consume any explicit labeled data and generates high quality paraphrases of the agent response by identifying relevant keywords in the agent response. For the purpose of the experiments in this paper, we consider part-of-speech (POS) tagging~\footnote[3]{http://www.nltk.org/book/ch05.html} to identify the keywords. Concretely, for each actual agent response, we extract the nouns and the verbs and treat them as the list of keywords. We address this list as the ``Control Words''. Table~\ref{table:control-word-samples} shows a few samples of control words extracted from the agent response.

\begin{table}[!htbp]
\centering
\begin{tabular}{|m{3.8cm}|m{3.2cm}|} \hline
\textbf{Actual Agent Response} & \textbf{Control Words} \\ \hline
I reqeust you to please wait for the same on this and the order will be confirmed shortly by the restaurant. & 'restaurant', 'reqeust', 'order', 'confirmed', 'please' \\ \hline
The food will be delivered as promised, our partner is on it the earleist. & 'food', 'partner', 'earleist', 'delivered' \\ \hline
Upendra, I will definitely help you. & 'help', 'Upendra' \\ \hline
I just have a word with the restaurant and they are closed for today so shall I proceed for the cancelation? & 'proceed', 'closed', 'restaurant', 'cancelation', 'today', 'word' \\ \hline
\end{tabular}
\caption{A few samples of actual agent response and the corresponding Control Words extracted using POS tags.}
\label{table:control-word-samples}
\end{table}

Given the context of the conversation and the control words, we fine-tune the GPT-2 model to generate candidate paraphrases for each agent response.  



\subsection{Semantic Similarity}\label{methodology-semantic-similarity}
Manually evaluating the semantic similarity of paraphrases generated is very expensive. To address this, we use BERTScore~\cite{bert-score} to capture the semantic similarity between the generated paraphrases and the actual agent responses. Concretely, BERTScore computes the cosine similarity for each token in the generated paraphrase with each token in the actual agent response using contextual embeddings.

\subsection{Textual Entailment}
One of the distinctive and critical requirements of our paraphrase generation problem setting is the necessity to match the context of the conversation with the generated paraphrase of the agent response. We therefore need to ensure that the context entails the generated paraphrase. To assess textual entailment, we train a model to generate a joint space embedding for conversation contexts and agent responses by using a ranking based loss~\cite{neculoiu-etal-2016-learning, DBLP:journals/corr/abs-1908-10084}. The model is trained to generate similar embeddings for a given context and the agent response that followed the context. We hypothesize that these embeddings can additionally capture similarity between the context and agent responses that didn't truly follow the given context but are similar to the actual agent response. 

Traditionally, generating such embeddings demands the availability of not only positive but also negative training samples where the response does not entail the context at hand. Given a domain specific problem setting and unavailability of labeled training data, the training data only encompasses positive samples where an agent response has truly followed conversation context. To identify negative samples, we leverage Multiple Negatives Ranking loss~\cite{henderson2017efficient} where negative samples are identified for a given positive training sample by treating the remaining samples in given training batch as negative samples. This approach works in our setting as the number of possible irrelevant responses for a given context tends to be higher than the number of possible relevant responses. 

For the purpose of the experiments in the paper, we employ the S-RoBERTa model~\cite{reimers2019sentencebert}. First, we validate the performance of the approach on the Ubuntu V2 Dialogue Corpus~\cite{ubuntu-corpus-2017} benchmark dataset. To reproduce the problem setting, the model is trained using Multiple Negatives Ranking loss by only considering the positive pairs in the training data and then evaluated on Next Utterance Classification (NUC)~\cite{ubuntu-corpus-2017} using nine false examples. Tables~\ref{table:ubuntu-datasize} and ~\ref{table:ubuntu-performance} shows the details of the dataset and the performance of the model. The approach is then reproduced on customer-agent chat conversations to capture the textual entailment of the responses given the context. 

\begin{table}[!htbp]
\centering
\begin{tabular}{|m{3.8cm}|m{2cm}|} \hline
 \textbf{Dataset} & \textbf{Size} \\ \hline
Train (Only Positive) & 49,916 \\ \hline
Validation & 18,565 \\ \hline
\end{tabular}
\caption{Dataset size for Ubuntu V2 Dialogue Corpus Train and Validation}
\label{table:ubuntu-datasize}
\end{table}

\begin{table}[!htbp]
\centering
\begin{tabular}{|m{3.8cm}|m{2cm}|} \hline
 \textbf{Metric} & \textbf{Score} \\ \hline
1 in 10 R@1 & 0.548 \\ \hline
1 in 10 R@2 & 0.700 \\ \hline
Mean Reciprocal Rank & 0.693 \\ \hline
\end{tabular}
\caption{Model performance on Ubuntu V2 Dialogue Corpus using 9 false examples }
\label{table:ubuntu-performance}
\end{table}

\subsection{Expression Diversity}
Our definition of quality of paraphrases generated also depends on the expression diversity a.k.a lexical dissimilarity between the generated response and the actual response. We apply $inverseBLEU$, a variation of the BLEU metric~\cite{bleu}, to quantify the expression diversity of the response. Let $r$ and $r*$ represent the generated response and the actual response. Eq~\ref{expression-diversity} describes how we calculate inverseBLEU.
\begin{equation}
\label{expression-diversity}
    inverseBLEU(r, r*) = 1 - BLEU(r, r*)
\end{equation}

In our setup, the expression diversity of the generated paraphrase is equal to the $inverseBLEU$ score. 

\subsection{Fluency}\label{methodology-fluency}
One additional metric that paraphrases should account for is language fluency. For the purpose of the experiments in this paper, we train our own model using the Corpus of Linguistic Acceptability (CoLA)~\cite{warstadt2019neural} dataset to score the grammatical correctness of the generated paraphrases. We consider ALBERT-base-v2 and RoBERTa-base for the purpose of the experiments and observe that RoBERTa performs better than ALBERT. Tables~\ref{table:cola-datasize} and \ref{table:cola-performance} describes the dataset and the model performances using Matthew's Correlation Coefficient (MCC)~\cite{warstadt2019neural}.

\begin{table}[!htbp]
\centering
\begin{tabular}{|m{2cm}|m{2.5cm}|m{1.5cm}|} \hline
 \textbf{Dataset} & \textbf{Domain} & \textbf{Size} \\ \hline
 Train & \hfil - & 8,551 \\\hline
\multirow{2}{\linewidth}{Dev} & In-domain & 527 \\ \cline{2-3}
 & Out-of-domain & 516 \\ \hline
\multirow{2}{\linewidth}{Test} & In-domain &  530 \\ \cline{2-3}
 & Out-of-domain & 533 \\ \hline
\end{tabular}
\caption{Dataset size for CoLA Train and Validation}
\label{table:cola-datasize}
\end{table}

\begin{table}[!htbp]
\centering
\begin{tabular}{|m{1.1cm}|m{1.8cm}|m{1.5cm}|m{1.5cm}|} \hline
 \textbf{Dataset} & \textbf{Domain} & \multicolumn{2}{|m{3cm}|}{\hfil \textbf{Model}}\\ \cline{3-4}
  & & ALBERT & RoBERTa \\ \hline
 
\multirow{2}{\linewidth}{Dev} & In-domain & 0.47 & 0.67 \\ \cline{2-4}
 & Out-of-domain & 0.45 & 0.54 \\ \hline
\multirow{2}{\linewidth}{Test} & In-domain & 0.5 & 0.63 \\ \cline{2-4}
 & Out-of-domain & 0.47 & 0.56 \\ \hline
\end{tabular}
\caption{Performances of ALBERT-base-v2 and RoBERTa-base on CoLA Dev and Test measured using MCC}
\label{table:cola-performance}
\end{table}


\subsection{Fine-tuning with RL}

Reward learning has enabled the success of Deep Reinforcement Learning (DRL) in a variety of NLI  tasks~\cite{ziegler2020finetuning, ranzato2015sequence, wu2018study}. We leverage a similar framework to further improve the quality of the generated paraphrases.

Let $X$ and $Y$ represent the input text and the generated text. Eq~\ref{rl-methodology} describes  trained language model P. 
\begin{equation}\label{rl-methodology}
    P(Y_{1:T}|X) = \prod_{t=1}^{T}  P(Y_{t}:Y_{1:t-1}, X)
\end{equation}

Inspired by previous success~\cite{DBLP:journals/corr/abs-1711-00279}, we design a generator-evaluator based RL framework with the difference that our evaluator is a pre-trained model. For the purpose of our experiments, we adopt the GPT-2 model as the generator and a model that generates a composite score by taking the average of the individual metrics defined in Section~\ref{methodology-semantic-similarity} through Section~\ref{methodology-fluency} as the evaluator. Concretely, we initialize the generator $G$ with policy $\pi = P$ and fine tune $G$ using RL. 
Let $X$, $Y$ and $\hat{Y}$ denote the context, the actual agent response and the generated paraphrase respectively. The evaluator $R(X, Y, \hat{Y})$ returns a normalised score between 0 and 1. 

Inspired by the success of framework presented in~\cite{ziegler2020finetuning}, we optimize the evaluator (reward function) as shown in Eq~\ref{Extending the approach to Reinforcement Learning} and Eq~\ref{modified-reward}.

\begin{align}
     E_{\pi}[r] &= E_{(X,Y) \sim D, \hat{Y}\sim\pi(\cdot|X)}[R(X, Y,\hat{Y})]\label{Extending the approach to Reinforcement Learning}\\
     R^{'}(X,Y,\hat{Y}) &= R(X,Y,\hat{Y}) - \beta\log\frac{\pi(Y/X)}{P(Y/X)}\label{modified-reward}
\end{align}
\\
We train the policy using the Proximal Policy Optimization (PPO)~\cite{schulman2017proximal} algorithm.
\section{Experiments}
\label{Experiments}
In this section, we describe the datasets, different experiment Setting and the results based on our automated metrics.

\subsection{Dataset}
The dataset considered for the purpose of the experiments comprises of 70,000 training samples and 5,000 validation samples selected randomly from 54,034 conversations. Each sample comprises of the conversation context, the control words and the actual agent response. The context is constructed using the 6 most recent messages exchanged between the customer and the agent which is later truncated based on the lengths of the tokenized contexts to ensure the model is able to consume the entire input. The control words are extracted as described in Section~\ref{lexical-control-methodology} and the response is the message used by the agent to respond to the customer in that context.

The custom model trained for textual entailment comprised of 110,721 samples selected randomly from 68,539 conversations.

\subsection{Experiment Setting}
For the purpose of the experiments, we leverage GPT-2 small with 117M parameters. The context and the control words are concatenated and passed with a context-control separator token  while the individual words in the list of control words are separated by control words separator token. The control words are passed in a random order to increase robustness of the model. 

\subsection{Baselines}
Table~\ref{table:model-performance} shows the performances of the different baselines across different metrics. 
To validate the effectiveness of our proposed approach, we consider the following baselines approaches using GPT-2

\begin{itemize}
  \item \textbf{Pretrained:} We consider the pre-trained GPT-2 model off-the-shelf to generate candidate paraphrases without any fine-tuning.
  \item \textbf{Only Context:} We fine-tune the GPT-2 model to generate the paraphrases of the agent response using  only the context data.    
  \item \textbf{Only Control Words:} The model is fine-tuned to generate the paraphrases using only the control words extracted from each agent response  
  \item \textbf{Context and Control Words:} We fine-tune the model to generate paraphrases using both the context and the control words 
  \item \textbf{Context and Control Words with Noise:} This setting is similar to ``Context and Control Words'' with the difference that 50\% of the words in the list of control words are randomly chosen and replaced with words chosen from the training dataset vocabulary at random.
  \item \textbf{Context and Control Words with Sampling:} This setting is similar to ``Context and Control Words'' with the difference that the control words are reduced with random sampling. The sampling rate is varied between 0.5 to 1. 
\end{itemize}


\begin{table*}[ht]
\centering
 \begin{tabular}{|p{4.7cm}| p{1.7cm}| p{1.7cm} |p{1.7cm}|p{1.5cm}|p{1.5cm}|} 
 \hline
 \textbf{Approach} & \multicolumn{5}{|m{6.4cm}|}{ \textbf{Scores}} \\\cline{2-6}
  & \textbf{Semantic Similarity} & \textbf{Textual Entailment} & \textbf{Expression Diversity} & \textbf{Fluency} & \textbf{Overall Score} \\ 
 \hline
Pre-trained & 0.268 & 0.559 & 0.954 & 0.859 & 0.660   \\ 
 \hline
Only Context &  0.632 & 0.695 & 0.762 & 0.886 & 0.744  \\ 
 \hline
Only Control words & 0.821 & 0.748 & 0.362 & 0.825 & 0.689  \\ 
 \hline
 Context and  Control words &  0.812 & 0.752 & 0.390 & 0.829 & 0.696  \\ 
 \hline
Context and Control words with Noise & 0.687 & 0.695 & 0.589 & 0.729 & 0.675  \\ 
 \hline
Context and Control words with Sampling &  0.732 & 0.731 & 0.560 & 0.867 & 0.722   \\ 
 \hline
\textbf{RL based Finetuning} &  0.671 & 0.726 & 0.742 & 0.873 & \textbf{0.753}  \\ 
 \hline
\end{tabular}
\caption{Performance metrics across various modelling approaches using GPT-2.}\label{table:model-performance}
\end{table*}

\begin{table*}[ht]
\centering
 \begin{tabular}{|p{4.0cm}| p{1.8cm}| p{1.8cm} |p{1.8cm}|p{1.8cm}|p{1.8cm}|} 
 \hline
 \textbf{Approach} & \multicolumn{5}{|m{6.4cm}|}{ \textbf{Scores}} \\\cline{2-6}
  & \textbf{Semantic Similarity} & \textbf{Textual Entailment} & \textbf{Expression Diversity} & \textbf{Fluency} & \textbf{Overall Score} \\ 
 \hline
UPSA & 1.84 $\pm$ 0.57 & 1.86 $\pm$ 0.48 & 3.79 $\pm$ 1.00 & 2.57 $\pm$ 0.77 & 2.52 $\pm$ 0.36   \\ 
 \hline
Only Context & 2.44 $\pm$ 0.96 & 3.38 $\pm$ 1.09 & 3.83 $\pm$ 1.01 & 4.60 $\pm$ 0.63 & 3.56 $\pm$ 0.46  \\ 
 \hline
Context and Control words with Sampling &  3.55 $\pm$ 1.22 & 3.88 $\pm$ 1.05 & 3.37 $\pm$ 1.03 & 4.60 $\pm$ 0.68 & 3.85 $\pm$ 0.55   \\ 
 \hline
\textbf{RL based Finetuning} &  3.74 $\pm$ 1.30 & 4.00 $\pm$ 1.06 & 3.83 $\pm$ 0.90 & 4.44 $\pm$ 0.88
 & \textbf{4.00} $\pm$ \textbf{0.66} \\ 
 \hline
\end{tabular}
\caption{Human Studies across various modelling approaches using GPT-2.}\label{table:model-performance-human}
\end{table*}

\subsection{RL Fine-tuning Settings}


The RL policy is initialized with the baseline ``Context and Control Words with sampling''. 
We replicate the fine-tuning settings applied in \cite{ziegler2020finetuning}. We fine-tuned the model for 400 steps with batch size 8 and 4 PPO epochs per batch with one minibatch each, and default values for the other parameters. We train the model on Tesla V100-SXM2-16GB GPU for 2 hours.

\subsection{Results and Analysis}
Tables~\ref{table:model-performance} and ~\ref{table:model-paraphrase-generation-sample} shows the performance of different approaches with the best approach highlighted in bold and paraphrase generations of the corresponding approaches for a sample conversation context and response. 

\begin{table*}[!htbp]
\centering
 \begin{tabular}{|p{4.7cm}| p{2.7cm}| p{2.7cm} |p{4.7cm}|} \hline
 \textbf{Context} & \textbf{Actual Agent Response} & \textbf{Approach} & \textbf{Generated Response} \\ 
 \hline
 \multirow{6}{\linewidth}{\hfil \textbf{agent} >> shall i transfer the chat to our senior management team? \\
 \textbf{customer} >> call me \\
 \textbf{customer} >> i can tell you how big liars are your delivery partners are \\
 \textbf{agent} >> i''m so sorry we don;t have the option now to callback you.} & \multirow{6}{\linewidth}{\hfil I can transfer the chat to our senior escallation desk, shall I transfer?.} & Only Context &  Please make sure to connect to our senior management team at 982****957 to check on the same. \\ \cline{3-4}
  & & Only control words & I can transfer this chat to the desk and you can assign a new Escallation or you can assign a different desk for me \\ \cline{3-4}
  & & Context and Control Words & I can transfer the Escallation Desk chat to the senior desk and they will be immediately reallaying. \\ \cline{3-4}
  & & Context and Control Words with Noise & We can only cancel the delay with an oepoe of coupon. \\ \cline{3-4}
 & & Context and Control Words with Sampling & Shall I transfer this chat to our senior management team?  \\ \cline{3-4}
  & & RL Finetuning & Should I transfer the chat to our senior management team? \\ \hline
 
 \end{tabular}
 \caption{Paraphrase generations using various approaches for a sample Conversation Context and Response}\label{table:model-paraphrase-generation-sample}
\end{table*}

The ``Pre-trained'' approach leverages the model trained on benchmark datasets in English~\cite{warstadt2019neural} to generate highly fluent English sentences. Hence, the fluency score in this case is high. However, the generated responses are highly irrelevant to the context and the actual agent response, thus, resulting in a poor Semantic Similarity and Textual Entailment score with high Diversity. ``Only Context'' achieves a higher score in textual entailment and expression diversity. This behaviour is expected as the model is fine-tuned using the contexts of the conversation. However, the semantic similarity is low as the actual agent response can still be different from the generated response. Concretely, our problem setting involves conversation contexts with multiple valid agent responses that can be semantically different due to untracked external signals. As a result, we observe ``Only Control words'' achieves a higher semantic similarity than ``Only Context''. But the overall score is still lower than ``Only Context'' as control words create an inherent bias in the model to generate lexically similar responses, and hence generate lower scores for expression diversity. ``Context and Control Words'' leverages both context and control words, and thus achieves a higher score overall. But the expression diversity is still low. 

To analyze the impact of the inclusion of control words, we run experiments with ``Context and Control Words with Noise'' and ``Context and Control words with Sampling''. We observe that adding noise in the form of random words along with the control words improves the expression diversity but generates relatively afluent and irrelevant responses, thus reducing the remaining scores. However, consuming randomly sampled words from the list of control words for inclusion in training performs better than adding noise to the control words with relatively higher expression diversity.

Based on the overall scores, we observe that ``Only Context'' achieves the highest performance followed by ``Control and Control words with Sampling''. For the purpose of our experiments, we still decided to choose ``Control and Control words with Sampling'' as ``Only Context'' had a higher performance mainly because of expression diversity while the former achieves relatively higher performance in both semantic similarity and textual entailment. ``RL based fine-tuning'' achieves a balanced score across different metrics with the highest overall score of 0.753. We verify the statistical significance of the result to observe that ``RL based fine-tuning''$(M = 0.753 ,SD = 0.095)$ significantly outperforms the second best performing approach (in overall score) ``Only Context'' $(M = 0.744,SD = 0.073), t(5000) = 5.33, p = 9.3\mathrm{e}{-8}$.

\subsection{Human Evaluation}
To further validate the quality of the paraphrases generated, we conducted a human evaluation study. Due to constraints on resources, 200 samples were selected at random for the evaluation. The samples were labelled on a scale of 1 to 5 (1 representing the worst and 5 representing the best performance) for each of “Semantic Similarity”, “Textual Entailment”, ”Expression Diversity” and ”Fluency”. Table~\ref{table:model-performance-human} shows the performance of different approaches validated by two annotators with subject matter expertise. To incorporate an existing benchmark for unsupervised paraphrase generation, we considered one additional baseline Unsupervised Paraphrasing by Simulated Annealing (UPSA)~\cite{liu-etal-2020-unsupervised} for the human evaluation. Additionally, we consider ”Only Context”, “Context and Control words with Sampled” and “RL based finetuning” approaches.  We observed that scores generated by our automated metric agree with the labels provided by the annotators.






\section{Conclusion}
\label{Conclusion}
In this paper, we considered the use case of paraphrase generation for agent responses in customer-agent chat conversations and demonstrated that RL based fine-tuning of GPT-2 can be leveraged to generated candidate paraphrases. We proposed applying lexical control to agent responses to enable unsupervised paraphrase generation and combined it with the conversation context to further improve the quality of paraphrases. We also explored automated paraphrase evaluation with custom metrics to measure various desired qualities of a paraphrase in a customer-agent chat conversation and then combined the metrics to construct the reward for RL fine-tuning. We show that the RL based approach significantly outperforms the rest of the approaches based on the automated metric scores. In future work, we plan to investigate broader implications of the proposed framework on similar problem statements with a wider range of external signals to control the text generation.

\section{Ethical Considerations}
 The proposed framework has been built to assist agents with paraphrased responses to improve customer experience and engagement. The data does not use any customer specific information, and hence, does not introduce bias specific to the customers.

\bibliography{anthology,naacl2021}
\bibliographystyle{acl_natbib}

\clearpage

\end{document}